\documentclass[10pt,twocolumn,letterpaper]{article}

\usepackage{cvpr}
\usepackage{times}
\usepackage{epsfig}
\usepackage{graphicx}
\usepackage{amsmath}
\usepackage{amssymb}
\usepackage{multirow}
\usepackage{bm}
\usepackage{marvosym}
\usepackage{booktabs}
\usepackage[accsupp]{axessibility} 

\begin{document}

\title{A Multi-Stage Adaptive Feature Fusion Neural Network for Multimodal Gait Recognition}

\author{Shinan Zou$^1$$^*$  \qquad Jianbo Xiong$^1$$^*$ \qquad Chao Fan$^2$$^,$$^3$ \qquad Shiqi Yu$^2$$^,$$^3$ \qquad Jin Tang$^1$$^\dagger$  \\
	$^1$School of Automation, Central South University \\
	$^2$Department of Computer Science and Engineering, Southern University of Science and Technology\\
	$^3$Research Institute of Trustworthy Autonomous System, Southern University of Science and Technology\\
	{\tt\small \{zoushinan,jianbo\_x,tjin\}@csu.edu.cn,12131100@mail.sustech.edu.cn,yusq@sustech.edu.cn}
}

\maketitle
\thispagestyle{empty}

\renewcommand{\thefootnote}{}
\footnotetext{$^*$ Equal contribution. }
\begin{abstract}
Gait recognition is a biometric technology that has received extensive attention. Most existing gait recognition algorithms are unimodal, and a few multimodal gait recognition algorithms perform multimodal fusion only once. None of these algorithms may fully exploit the complementary advantages of the multiple modalities. In this paper, by considering the temporal and spatial characteristics of gait data, we propose a multi-stage feature fusion strategy (\textbf{MSFFS}), which performs multimodal fusions at different stages in the feature extraction process. Also, we propose an adaptive feature fusion module (\textbf{AFFM}) that considers the semantic association between silhouettes and skeletons. The fusion process fuses different silhouette areas with their more related skeleton joints. Since visual appearance changes and time passage co-occur in a gait period, we propose a multiscale spatial-temporal feature extractor (\textbf{MSSTFE}) to learn the spatial-temporal linkage features thoroughly. Specifically, MSSTFE extracts and aggregates spatial-temporal linkages information at different spatial scales. Combining the strategy and modules mentioned above, we propose a multi-stage adaptive feature fusion (\textbf{MSAFF}) neural network, which shows state-of-the-art performance in many experiments on three datasets. Besides, MSAFF is equipped with feature dimensional pooling (\textbf{FD Pooling}), which can significantly reduce the dimension of the gait representations without hindering the accuracy. https://github.com/ShinanZou/MSAFF
\end{abstract}

\section{Introduction}\label{sec1}
Unlike recognition methods that use fingerprint, face, and iris recognition, gait recognition does not require the subject's cooperation. It has the characteristics of non-contact, long-distance recognition. Thus, it can be widely applied in many fields, e.g., video surveillance and security system. However, there are still many problems in the practical application because many factors will significantly change the subject's appearance, such as clothing and cross-view.

In order to address the difficulties mentioned above, many multimodal gait recognition methods \cite{56,57,58,59,60,61} have been tried, which shows the feasibility of multimodal gait recognition. The commonly-used modalities for gait recognition are optical flow, silhouette, skeleton, inertial sensor data, etc. Our study chooses silhouettes and skeletons for the following reasons: 1) Data acquisition is easy. Silhouettes can be extracted from RGB images, and the skeletons can be obtained from gait videos using the human pose estimation algorithms \cite{63,64,65}, 2) The usage scenario is extensive because surveillance cameras are widely installed, and 3) The two data modalities complement each other. The silhouette is more sensitive to the change in human shape and can better capture the changes in human shape. However, in the case of self-occlusion, the silhouette will lose some information about the part of the human body. The skeleton can solve the problem of self-occlusion more effectively, but the changes in the human body during walking are usually ignored.

\begin{figure}[t]%
	\centering
	\includegraphics[width=1.0\linewidth, height=2.0cm]{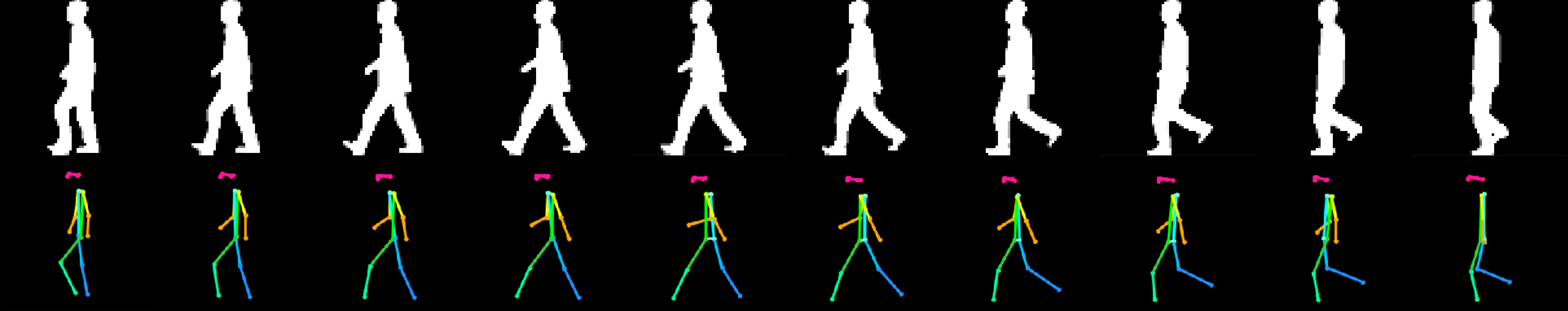}
	\caption{The first row is the silhouette sequence, and the second row is the skeleton sequence; they are the same subject from the CASIA-B dataset.\footnotesize}
		\vspace{-12pt}
	\label{Fig1}
\end{figure}
As shown in Figure \ref{Fig1}, silhouettes and skeletons are sequences. For this type of data, spatial features and temporal features are essential to represent them. We note that some unimodal methods \cite{6,9,10,66,67,68} extract spatial and temporal features in different ways and achieve state-of-the-art (SOTA) results on publicly available gait datasets. However, the existing multimodal methods \cite{56,57,58,59,60,61} perform only one fusion; they ignore the spatial and temporal characteristics of gait data in the fusion stage; the spatial, temporal, and spatial-temporal features are essential for a complete representation of the gait, so it is necessary to fuse the spatial, temporal, and spatial-temporal characteristics of the different modalities separately to exploit the advantages of modal complementarity further. We propose a multi-stage feature fusion strategy (MSFFS) to solve this problem. The three stages are as follows: 1) Frame level fusion, i.e., frame-to-frame fusion of silhouette and skeleton features to obtain the spatially complementary features of a single frame; 2) Spatial-temporal level fusion, i.e., the spatial-temporal features of two modalities are fused to obtain spatial-temporal complementary information; and 3) Global level fusion. To better represent gait, silhouette features, skeleton features, frame level fusion, and spatial-temporal level fusion features are fused again.

\begin{figure}[t]%
	\centering
	\includegraphics[width=0.8\linewidth,height=2.5cm]{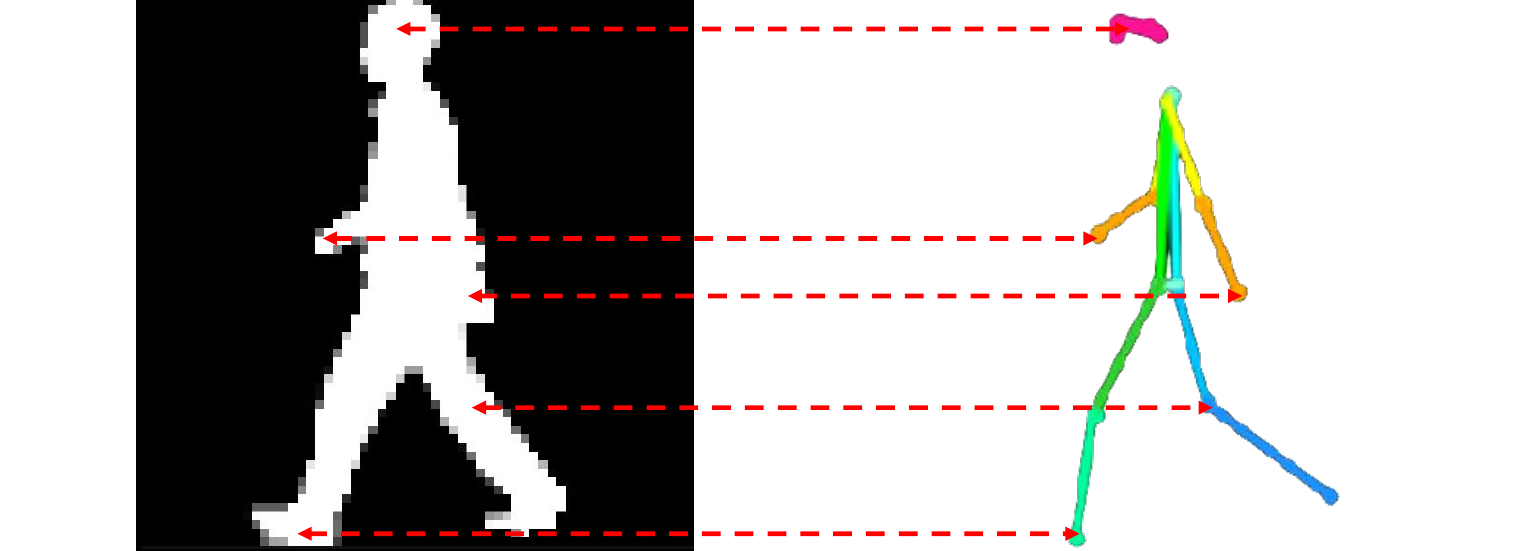}
	\caption{Each part of the silhouette corresponds to different joints of the skeleton.\footnotesize}
		\vspace{-12pt}
	\label{Fig2}
\end{figure}
Furthermore, as shown in Figure \ref{Fig2}. The semantic correlation between the two data modalities is represented spatially as the correspondence between the parts of the silhouette and the skeleton joint positions. In order to fully explore the relevance of semantic information, we pay more attention to the corresponding parts of silhouettes and skeletons while paying less attention to the parts that do not correspond. This point of view has not been mentioned in previous work. Thus, we propose an adaptive feature fusion module (AFFM), and different silhouette areas will be fused with more related skeleton joints.

In terms of the sequence of processing spatial information and temporal information, some existing methods can be divided into four types: 1) Extracting temporal features first, spatial features later, e.g., Gait Energy Image (GEI) \cite{1}; 2) Extracting spatial features first and followed by the temporal features, e.g., GaitSet \cite{9}, GaitPart \cite{10}; 3) Extracting spatial features first, temporal features later, then the spatial and temporal features are extracted independently, e.g., CSTL \cite{66}; 4) Extracting spatial and temporal features simultaneously, e.g., MT3D \cite{67}, GaitGL \cite{68}. Due to the significant separation between temporal and spatial feature extraction, The first three methods can potentially hinder spatial-temporal information interactions. The fourth method is based on 3D-CNN. It can extract temporal and spatial features simultaneously but must be stacked many times to obtain the global-wise field. So this method may bring problems of parameter redundancy and massive resource consumption. In addition, we note that GaitSet \cite{9} only focuses on the whole body shape when extracting features; this approach will lose fine-grained information at different local scales and affect the gait recognition results. GaitPart \cite{10} learn the spatial and temporal representation of different parts independently. However, it ignores that all human body parts are interconnected and move jointly during walking.

To address the above issues, we propose a novel module capable of extracting spatial-temporal linkage features simultaneously at different spatial scales, named multiscale spatial-temporal feature extractor (MSSTFE). MSSTFE extracts spatial-temporal linkage features in three spatial scales, i.e., part level, local level, and global level. In this way, fine-grained information about the part and local areas can be preserved, while the linked features of the global area can also be obtained. We use a stacking structure that combines spatial one-dimension (1D) convolution and temporal 1D convolution when extracting spatial-temporal linkage features, which ensures a low training computational cost, and the extraction processes of spatial and temporal features are not significantly separated. In addition, we find that different parts of the human body have different motion amplitudes. Accordingly, we use the attention mechanism to capture the parts with greater motion amplitudes in the global area to obtain a more accurate gait representation.  

The MSFFS and MSSTFE significantly increase the dimension of the gait representations, increasing the recognition process's computational complexity. We propose a simple but efficient method, feature dimensional pooling (FD Pooling), to solve this problem. FD Pooling can significantly reduce the dimension of the gait representations with almost no loss of accuracy.

The main contributions are summarized as follows:

$\bullet$ We present a multi-stage feature fusion strategy (MSFFS) that fully exploits the complementary advantages of silhouettes and skeletons.

$\bullet$ We propose an adaptive feature fusion module (AFFM) to capture the relational semantic information of silhouettes and skeletons.

$\bullet$ We design an efficient spatial-temporal feature extraction module: multiscale spatial-temporal feature extractor (MSSTFE), which extracts the spatial-temporal linkage feature on different spatial scales.

$\bullet$ We propose feature dimensional pooling (FD Pooling) to reduce the dimension of the gait representations simply and efficiently.

$\bullet$ By combining the abovementioned strategies and modules, we propose a multi-stage adaptive feature fusion neural network (MSAFF) for multimodal gait recognition. Our network outperforms the state-of-the-art on CASIA-B \cite{25}, Gait3D \cite{75} and GREW \cite{77} gait datasets.
\begin{figure*}[t]%
	\centering
	\includegraphics[width=0.9\textwidth, height=6cm]{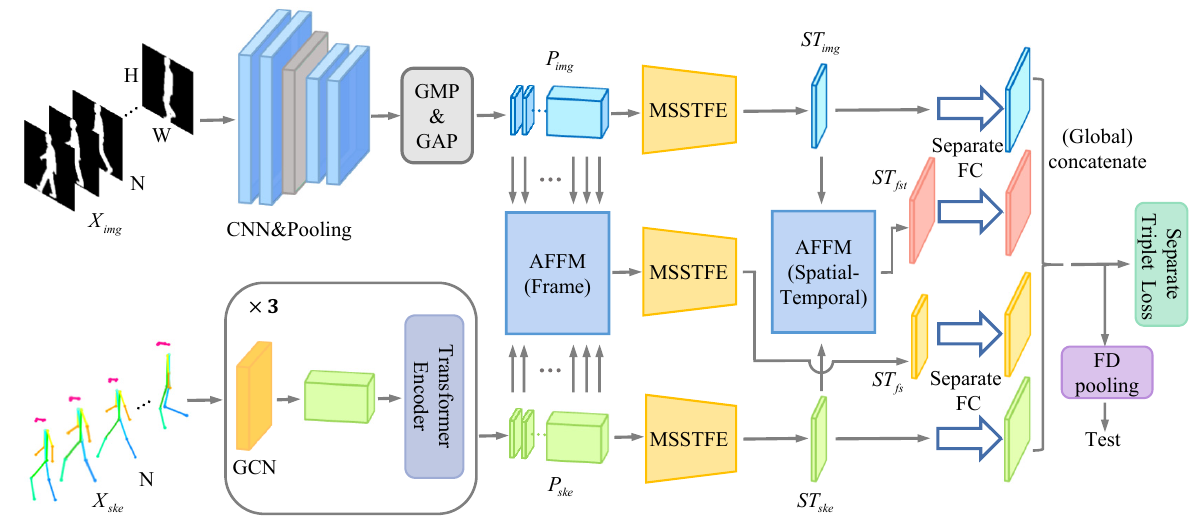}
	\caption{Overview of the proposed network.\footnotesize}
		\vspace{-12pt}
	\label{Fig3}
\end{figure*}
\section{Related Work}\label{sec2}
\subsection{Unimodal Gait Recognition}
Unimodal gait recognition methods can be divided into model-based and appearance-based. Model-based methods \cite{31,62,32} model humans' structure and walking action by extracting human pose information \cite{63,64,65,82,83,84}. Moreover, model-based methods have strong robustness to occlusion and viewing angles but ignore the shape of human walking. However, the model-based methods rely on the accuracy of the human pose estimation algorithm. The appearance-based methods \cite{1,2,3,4,5,13,14,17,18,19,20,21,9,10,66,67,68,78,79,80,81} currently perform better than the model-based methods but are affected by self-occlusion and viewing angles. These works in \cite{1,2,3,4,5,13,14,17,18,19,20,21} combine multiple images into one image along the temporal dimension, but the association information of adjacent frames may be lost. The method in \cite{6} uses LSTM to process time information but retains unnecessary constraints of the temporal sequences \cite{9}. Gaitset \cite{9} proposes to take gait as a set and aggregate temporal dimension through statistical functions. It is simple and effective and has attracted wide attention. GaitPart \cite{10} uses temporal multi-scale features module for modeling the short term. A multi-scale temporal features extraction method and a temporal feature aggregation module are proposed in CSTL \cite{66}. GaitGL \cite{68} designs local and global 3D-CNN to extract local and global spatial features.

\subsection{Multimodal Gait Recognition}
Facing complex gait problems, some works \cite{56,57,58,59,60,61,76} attempt to use multimodal methods, such as GaitCode \cite{56} uses the self-coding network to fuse acceleration and ground contact force data. The work in \cite{57} fuses data from motion sensors and visible light cameras for gait recognition at the decision level. These methods use inertial sensor data, usually obtained by smart wearable devices or portable mobile phones, and the application scenarios may be limited. The method in \cite{58} trains multiple multimodal CNNs using optical flow, grey pixel, and depth information. Autoencoder GaitNet is proposed in \cite{59}, which automatically learns the unentangled gait features (posture and appearance) from videos. The multimodal gait recognition system in \cite{60} combines RGB, optical flow, and audio. UGaitNet \cite{61} addresses the problem of modal loss in multimodal frameworks. Transgait \cite{76} combines silhouettes and pose heatmaps to mine the gait feature of a pedestrian. The multimodal algorithms described above only perform one fusion without considering the gait's spatial, temporal, and spatial-temporal characteristics, and the advantages of multimodal complementarity are not fully exploited.

\section{Method}\label{sec3}
\subsection{Pipeline}\label{sec3.1}
The overview of the proposed gait recognition network is shown in Figure \ref{Fig3}. The input of our network is a silhouette sequence $\ X_{img}\in\mathbb{R}^{N\times H\times W}$ with frame number $N$, height $H$ and width $W$ and a skeleton sequence $X_{ske}\in\mathbb{R}^{N\times3\times Z}$ with frame number $N$, channel number $3$ and joints number $Z$, where channel number is the 2D coordinates of joints with confidence scores. The specific steps of our multimodal gait recognition network are as:
\begin{align}
\setlength{\abovedisplayskip}{6pt}
\setlength{\belowdisplayskip}{5pt}
\label{P1-P2}
{S}_{img}&=F({X}_{img}). \\
{S}_{ske}&=GT\left(X_{ske}\right). 
\end{align}
where $F$ and $GT$ are frame-level spatial feature extraction networks of silhouette and skeleton, respectively, $F$ contains four 2D convolution layers and a 2D max pooling layer, while $GT$ contains three graph convolution transformer layers, $S_{img}\in\mathbb{R}^{N\times C\times H/2\times W/2}$ and $S_{ske}\in\mathbb{R}^{N\times C\times Z}$, and $C$ refers to the channel number of the feature.

$S_{img}$ is divided into $K$ parts according to the silhouette from top to bottom, and then the Global Max Pooling (GMP) and Global Average Pooling (GAP) are applied to each part to obtain the part-level features $P_{img}\in\mathbb{R}^{N\times C\times K}=\{p_{img}^i\in\mathbb{R}^{C\times K}\vert i=1,\ldots,N\}$, i.e.,
\begin{equation}
\setlength{\abovedisplayskip}{6pt}
\setlength{\belowdisplayskip}{5pt}
\label{P3}
P_{img}=GAP\left(S_{img}\right)+GMP\left(S_{img}\right).
\end{equation}

For${\ S}_{ske}$, we treat a joint as a part and sequence the joints in coco format \cite{73}, i.e. 
\begin{equation}
\setlength{\abovedisplayskip}{6pt}
\setlength{\belowdisplayskip}{5pt}
\label{P4}
P_{ske}=CoCo(S_{ske}).
\end{equation}
where $P_{ske}\in\mathbb{R}^{N\times C\times Z}=\{p_{ske}^i\in\mathbb{R}^{C\times Z}\vert i=1,\ldots,N\}$.

The intermediate variables $P_{img}$ and $P_{ske}$ are then sent into two independent multiscale spatial-temporal feature extractors (MSSTFE) to obtain the spatial-temporal linkage features. i.e.,
\begin{align}
\setlength{\abovedisplayskip}{6pt}
\setlength{\belowdisplayskip}{5pt}
\label{P5-P6}
{ST}_{img}&={MSSTFE}_{img}({P}_{img}). \\
{ST}_{ske}&={MSSTFE}_{ske}({P}_{ske}). 
\end{align}
where, $ST_{img}\in\mathbb{R}^{C\times3K}$and $ST_{ske}\in\mathbb{R}^{C\times3Z}$.

Simultaneously with the previous step, $P_{img}$ and $P_{ske}$ are input to the adaptive feature fusion module (AFFM) for frame level fusion, and the output is $P_{fs}\in\mathbb{R}^{N\times C\times K}$. $P_{fs}$ is input into the third independent multiscale spatial-temporal feature extractor (MSSTFE) to obtain the spatial-temporal linkage feature $ST_{fs}\in\mathbb{R}^{C\times3K}$, i.e.\footnote{In Eq \ref{P7}, $Concat$ represents concatenation operations, when it appears again, it means the same operation.},
\begin{equation}
\setlength{\abovedisplayskip}{6pt}
\setlength{\belowdisplayskip}{5pt}
\label{P7}
	\begin{aligned}
P_{fs}&=Concat\left({AFFM}_1,...,{\rm AFFM}_N\right), \\
Where \quad  &{AFFM}_i=\ {{AFFM}_{frame}(p}_{img}^i, p_{ske}^i),
\end{aligned}
\end{equation}
\begin{equation}
\setlength{\abovedisplayskip}{6pt}
\setlength{\belowdisplayskip}{5pt}
\label{P8}
ST_{fs}={MSSTFE}_{fs}({P}_{fs}).
\end{equation}

Next, ${ST}_{img}$ and ${ST}_{ske}$ are input adaptive feature fusion modules (AFFM) which are used for multimodal fusion at spatial-temporal level, and the output ${ST}_{fst}\in\mathbb{R}^{C\times3K}$ is obtained, i.e.,
\begin{equation}\label{P9}
\setlength{\abovedisplayskip}{6pt}
\setlength{\belowdisplayskip}{5pt}
ST_{fst}={AFFM}_{st}({ST}_{img}, {ST}_{ske}).
\end{equation}

Next, several separate fully connected (FC) layers are used to map feature vectors $ST_{img}$, $ST_{ske}$, $ST_{fs}$ and ${ST}_{fst}$ to the metric space, next, global level fusion of the mapped feature vectors is performed to obtain the output $Out\in\mathbb{R}^{(9K+3Z)\times Out_C}$ and $Out_c$ is the channel number of FC. $Out$ is fed into the All (BA+) triplet loss \cite{28} for training the network. For efficiency during testing, $Out$ will be fed into feature dimensional pooling (FD Pooling) to reduce the dimension of the gait representations.

\subsection{Multi-Stage Feature Fusion Strategy}
\noindent\textbf{Description.} Multimodal feature fusion is used at frame level, spatial-temporal level, and global level.\par\noindent\textbf{Motivation.}  Spatial feature and temporal contain rich identification information mentioned in Section \ref{sec1}. In order to obtain the complementary advantages of silhouettes and skeletons, we perform multimodal fusion in multiple stages of feature extraction. At the same time, global level fusion makes the final measurement feature have multi-scale and multi-type information and obtain a more comprehensive feature representation.\par\noindent\textbf{Operation.}

\textbf{\emph{Frame level fusion.}} The $N$ silhouette frames and $N$ skeleton frames are frame-to-frame correspondences. The corresponding silhouette features and skeleton features will be paired into the adaptive feature fusion module (AFFM) for fusion during frame level fusion.

\textbf{\emph{Spatial-temporal level fusion.}} ${ST}_{img}$ and ${ST}_{ske}$ are obtained from silhouettes and skeletons after multiscale spatial-temporal linkage feature extraction. These two features are similar to the single frame features in terms of dimensionality. These two features are passed through the adaptive feature fusion module (AFFM) in pairs to obtain the spatial and temporal correlation semantic information. Here the AFFM used by the frame level and spatial-temporal level is parameter independent.

\textbf{\emph{Global level fusion.}} $ST_{img}$, $ST_{ske}$, $ST_{fs}$, and ${ST}_{fst}$ mentioned in Section \ref{sec3.1} will be spliced together after passing through their respective fully connected layers as the final gait representation for recognition.
\begin{figure}[t]%
	\centering
	\includegraphics[width=1.0\linewidth,height=3.7cm]{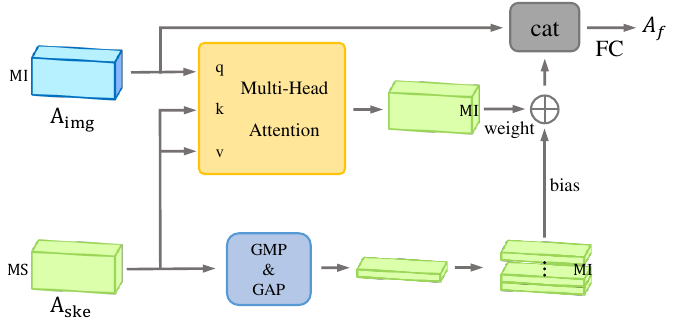}
	\caption{Overview of the adaptive feature fusion module (AFFM).\footnotesize}
	\vspace{-12pt}
	\label{Fig4}
\end{figure}
\subsection{Adaptive Feature Fusion Module}
\noindent\textbf{Description.} The adaptive feature fusion module (AFFM) adaptively searches for the associated weight between each part of the silhouette and skeleton feature so that the associated part has a higher weight; then, we fuse the silhouette and skeleton features according to the weight to get the fused feature.\par\noindent\textbf{Motivation.} Silhouette and skeleton are semantically related, and there is a corresponding relationship between the parts of the silhouette and the joint positions of the skeleton. Therefore, focusing on the fusion of related parts is more beneficial for obtaining the associated semantic information of the two data modalities.

\par\noindent\textbf{Operation.} The general process of operation is shown in Figure \ref{Fig4}. For the convenience of unified description, the input silhouette feature of AFFM is $A_{img}\in\mathbb{R}^{D\times M I}$, and the skeleton feature is $A_{ske}\in\mathbb{R}^{D\times M S}$, where $D$ denotes the number of feature channels, $MI$ and $MS$ represents the number of silhouette and skeleton feature parts, respectively. For the specific values of each parameter, $D=C$, $MI=K$, $MS=Z$ in frame level fusion. $D=C$, $MI=3K$, $MS=3Z$ in spatial-temporal level fusion, where $C$ refers to the number of channels, $K$ refers to the number of silhouette parts, and $Z$ refers to the number of skeleton joints. 

Next, the correlation between silhouette features and skeleton features computed is used as the weight matrix. According to the weight matrix, the features of different joints are weighted to obtain the weighted skeleton features. The weighted skeleton features will be calculated several times in parallel. Finally, the weighted skeleton features of each time will be spliced together and projected again. The process above is implemented through multi-head attention \cite{39}, i.e.,
\begin{equation}
\setlength{\abovedisplayskip}{6pt}
\setlength{\belowdisplayskip}{5pt}
\label{P10}
\begin{aligned}
&Weight_{ske}=Concat\left({Head}_1,...,{Head}_h\right)W^O,\\
where\\
{Head}_i&=softmax(\frac{{{(W_i^KA}_{img})}^T{(W_i^QA}_{ske})}{\sqrt d})\ {{(W_i^VA}_{ske})}^T.
\end{aligned}
\end{equation}

The parameter mapping matrix is: $W_i^K\in\mathbb{R}^{d\times D}$, $W_i^Q\in\mathbb{R}^{d\times D}$, $W_i^V\in\mathbb{R}^{d\times D}$, $W^O\in\mathbb{R}^{h\times d\times D}$, where $d=\frac{D}{h}/r$, $D$ is the number of feature channels mentioned above, $r$ stands for channel compression ratio and $h$ stands parallel attention layers or heads. We employ $h = 4$ in this work. 

Next, we use Global Max Pooling (GMP) and Global Average Pooling (GAP) to obtain the global spatial features of the skeleton, and the global spatial features of the skeleton are copied $MI$ times as the bias of the weighted skeleton features, i.e.,
\begin{equation}
\label{P11}
\setlength{\abovedisplayskip}{6pt}
\setlength{\belowdisplayskip}{5pt}
Bias_{ske} =Concat(G_1\ldots.G_{MI}),
\end{equation}
where $\quad G_i=GAP\left(A_{ske}\right)+GMP(A_{ske})$.

The final silhouette and skeleton fusion feature $A_f\in\mathbb{R}^{D\times M I}$ is 
\begin{equation}
\label{P12}
\setlength{\abovedisplayskip}{6pt}
\setlength{\belowdisplayskip}{5pt}
A_f=FC(Concat(A_{img}, Weight_{ske} +Bias_{ske})).
\end{equation}
where, $FC$ is a fully connected layer.

\subsection{Multiscale Spatial-Temporal Feature Extractor}
\noindent\textbf{Description.} The spatial-temporal linkage features are extracted from three spatial scales: part, local, and global.\par\noindent\textbf{Motivation.} The changes in the human body’s spatial shape and temporal motion co-occur for walking. Therefore, the extraction of spatial and temporal features should not be significantly separated, and the linkage between them contains rich identity features. As mentioned in Section 1, the movement of the human body with time can be observed at different spatial scales. The network can obtain the details of human spatial-temporal motion at different scales by extracting spatial-temporal linkage features at multiple spatial scales. Finally, the spatial-temporal linkage features of various scales are integrated to realize the communication and fusion of information at different scales. In addition, in the global space, each part of the human body has different degrees of motion and contributes to different gait identification characteristics. Thus, highlighting the parts that contribute more to gait identification features in the global space to obtain a more accurate representation of spatial and temporal linkage features.\par
\begin{figure}[t]%
	\centering
	\includegraphics[width=1.0\linewidth,height=6.6cm]{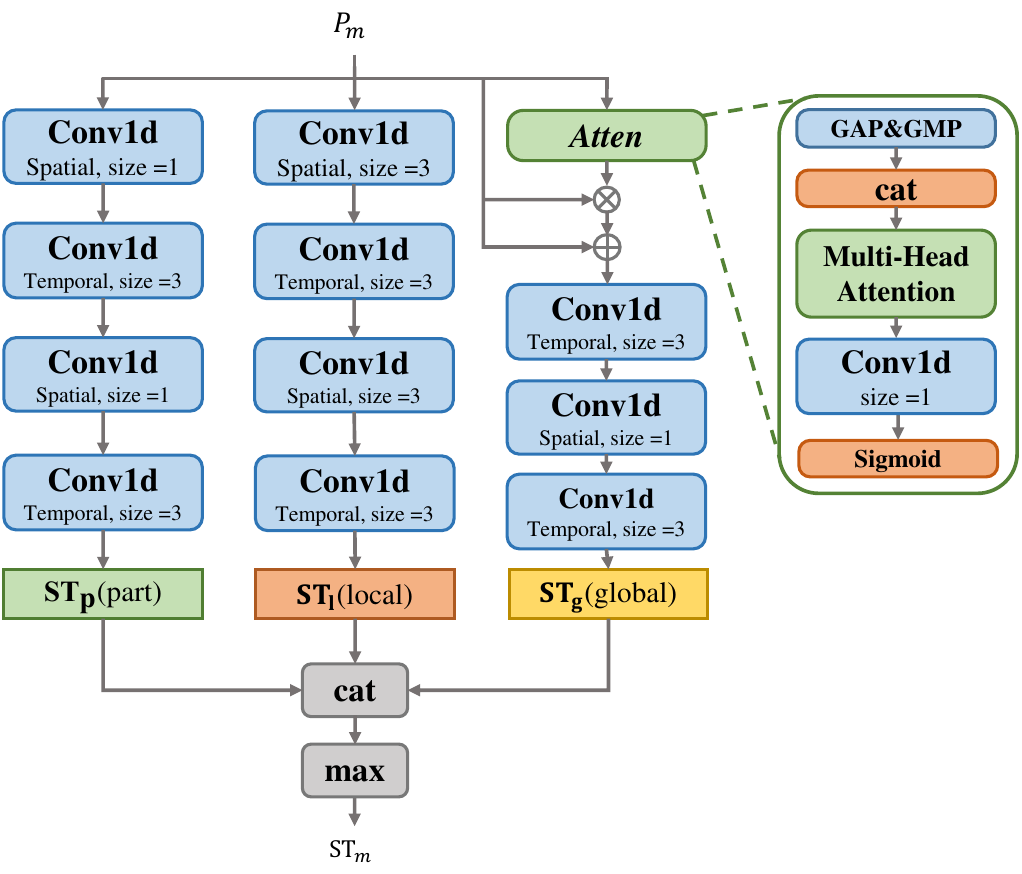}
	\caption{Overview of the multiscale spatial-temporal feature extractor (MSSTFE).\footnotesize}
	\label{Fig5}
		\vspace{-12pt}
\end{figure}

\noindent\textbf{Operation.} The specific details are shown in Figure \ref{Fig5}. For the convenience of unified explanation, the input of MSSTFE is $P_m\in\mathbb{R}^{N\times C\times M}$, and $M$ represents the number of feature parts. For specific parameters, $M=K$ when input is $P_{img}$ or $P_{fs}$, and $M=Z$ when input is $P_{ske}$. $K$ refers to the number of silhouette parts, and $Z$ refers to the number of skeleton joints. 

\textbf{\emph{Part level and local level.}} Firstly, a 1D convolution with a kernel size of $s_{size}$ is used to extract features along the spatial dimension, and then $M$ 1D convolution with a kernel size of $t_{size}$ is used to extract features along the temporal dimension of $M$ part features, respectively. In this way, the interaction of temporal and spatial dimensions is realized. It can be expressed as follows:
\begin{equation}
\setlength{\abovedisplayskip}{6pt}
\setlength{\belowdisplayskip}{5pt}
\label{P13}
P_{mst}={{Conv1d}_{mt1}(Conv1d}_{s1}\left(P_m\right)).
\end{equation}
where, $P_{mst}\in\mathbb{R}^{N\times C\times M}$.

Next, to expand the scope of spatial-temporal interaction, we repeat the above operations for $P_{mst}$ to complete the extraction of spatial-temporal linkage features. The operation is
\begin{equation}
\setlength{\abovedisplayskip}{6pt}
\setlength{\belowdisplayskip}{5pt}
\label{P14}
P_{mstst}={{Conv1d}_{mt2}(Conv1d}_{s2}\left(P_{mst}\right)),\\
\end{equation}
where, $P_{mstst}\in\mathbb{R}^{N\times C\times M}$.

The processes of both part level and local level can be described by the above two equations. The differences between them are: $ST_p = P_{mstst}$, $s_{size} = 1$, $t_{size} = 3$ when extracting the spatial-temporal linkage features on part level; $ST_l = P_{mstst}$, $s_{size} = 3$, $t_{size} = 3$ when extracting spatial-temporal linkage features on local level. $ST_p$ is Part level spatial-temporal linkage feature, and $ST_l$ is local level spatial-temporal linkage feature.

\textbf{\emph{Global level.}} We first compress the channel dimension of $P_m$ through Global Max Pooling and Global Average Pooling. Next, by splicing the compressed features to represent channel features. The process is
\begin{equation}
\setlength{\abovedisplayskip}{6pt}
\setlength{\belowdisplayskip}{5pt}
\label{P15}
POOL=Concat\left(GAP\left(P_m\right), GMP\left(P_m\right)\right),  
\end{equation}
where, $POOL=\{{pool}_i\in\mathbb{R}^{2\times M}\vert i=1\ldots.N\}$.

Then, the self-attention mechanism \cite{39} is used to capture the global spatial information of $POOL$, and the formula is expressed as:
\begin{equation}
\label{P16}
\setlength{\abovedisplayskip}{6pt}
\setlength{\belowdisplayskip}{5pt}
POOL\_A=Concat(A_1,...,A_N), 
\end{equation}
where, $A_i=softmax(\frac{{{(W_p^Kpool}_i)}^T{(W_p^Qpool}_i)}{\sqrt2}){{(W_p^Vpool}_i)}^T$.

The parameter mapping matrix is: $W_p^K\in\mathbb{R}^{2\times2}$, $W_p^Q\in\mathbb{R}^{2\times2}$, $W_p^V\in\mathbb{R}^{2\times2}$, $POOL\_A\in\mathbb{R}^{N\times2\times M}$.

Next, Global spatial attention is obtained by a 1D convolution with a convolution kernel size of 1 and sigmoid function, i.e.,
\begin{equation}
\setlength{\abovedisplayskip}{6pt}
\setlength{\belowdisplayskip}{5pt}
\label{P17}
Atten = sigmoid\left( Conv1d(POOL\_ A) \right).
\end{equation}

The global spatial attention weight is used to re-weight the spatial features, i.e.,
\begin{equation}
\setlength{\abovedisplayskip}{6pt}
\setlength{\belowdisplayskip}{5pt}
\label{P18}
P_{g} = Atten*~P_{m} + P_{m}.
\end{equation}

Then, $M$ 1D convolutions with a kernel size of 3 are used to extract features along the temporal dimension of $M$ part features, respectively, i.e.,
\begin{equation}
\setlength{\abovedisplayskip}{6pt}
\setlength{\belowdisplayskip}{5pt}
\label{P19}
P_{gt}={Conv1d}_{mt3}\left(P_g\right),
\end{equation}
where, $P_{gt}\in\mathbb{R}^{N\times C\times M}$.

In order to further expand the propagation range of global features along the temporal dimension, We reperform part-level spatial-temporal feature extraction on $P_{gt}$, similar to Equation \ref{P13}, i.e.,
\begin{equation}
\setlength{\abovedisplayskip}{6pt}
\setlength{\belowdisplayskip}{5pt}
\label{P20}
ST_g={{Conv1d}_{mt4}(Conv1d}_{s3}\left(P_{gt}\right)),
\end{equation}
where, $ST_g\in\mathbb{R}^{N\times C\times M}$.

Finally, we splice $ST_p$, $ST_l$, $ST_g$ together, then use the max pooling to aggregate the spliced features along the temporal dimension as the output of MSSTFE, i.e.,
\begin{equation}
\setlength{\abovedisplayskip}{6pt}
\setlength{\belowdisplayskip}{5pt}
\label{P21}
ST_m=MAX(Concat(ST_p, ST_l, ST_g)),
\end{equation}
where, $ST_m\in\mathbb{R}^{C\times3M}$.

\subsection{Feature Dimensional Pooling}
\noindent\textbf{Description.} Reducing the dimension of the gait representation under almost no loss of accuracy.\par\noindent\textbf{Motivation.} In this paper, the network output has a high representation dimension due to the use of MSFFS and MSSTFE. Therefore, the output dimension needs to be reduced to make the proposed network more suitable for practical application.\par\noindent\textbf{Operation.} First, we cut the output of network $Out$ into $op$ parts along the channel dimension, where the $i-th$ part is denoted as $out_i\in\mathbb{R}^{(9K+3Z)\times Out_C/op}$ and $Out_c$ is the output dimension. Next, we pool each part along the channel dimension, and connect all pooled parts, i.e.,
\begin{equation}
\setlength{\abovedisplayskip}{6pt}
\setlength{\belowdisplayskip}{5pt}
\label{P22}
Out_{FDP}=Concat(FDP_1\ldots.FDP_{op}),
\end{equation}
where, $FDP_i=Pooling(out_i)$.

In Equation \ref{P22}, $Out_{FDP}\in\mathbb{R}^{(9K+3Z)\times op}$ indicates that the output of feature dimensional pooling. $Pooling$ is a pooling operation. We try several pooling methods in the ablation experiment and find that the average pooling performed the best. We can control the output of feature dimensional pooling by controlling $op$.

\section{Experiments}\label{sec4}
\subsection{Datasets}
\textbf{CASIA-B \cite{25}} is the most widely-used gait dataset and comprises 124 subjects. For each subject, there are three gait conditions: normal walking (NM), bag walking (BG), and wearing a coat (CL). Each condition has 11 angles of view. The top 74 subjects are trained when evaluating performance, and the remaining 50 are retained for the test. The $NM\#1-NM\#4$ in the test phase are used as the gallery, and the remaining $NM\#5-NM\#6$, $BG\#1-BG\#2$ and $CL\#1-CL\#2$ are used as the probe.

\textbf{Gait3D \cite{75}} is a large 3D representation-based gait recognition dataset. It contains 4,000 subjects and over 25,000 sequences extracted from thirty-nine cameras in an unconstrained indoor scene. Subjects are divided into training and testing sets (3000 subjects are for training and 1000 for testing). In the test set,  one sequence is selected as the probe set, while the rest are regarded as the gallery set.

\textbf{GREW \cite{77}} is a large-scale and in-the-wild dataset. It includes 26,345 subjects and 128,671 sequences. There are 20,000 subjects for training, 6,000 for testing, and 345 for validating. GREW also contains a distractor set of 233,857 sequences (unlabeled data). In the testing set, there are 4 sequences in each subject, 2 for the query set and 2 for the gallery set.

\begin{table*}[ht]
	\renewcommand{\arraystretch}{1.2}
	\caption{Averaged rank-1 accuracies on CASIA-B, excluding identical-view cases.\footnotesize}
	\footnotesize
	\centering
	\label{tbl2}
	\begin{tabular}{cccccccccccccc}
		\hline
		\multicolumn{2}{c}{Gallery NM \#1-4}            & \multicolumn{11}{c}{Angle of view: $0^o-180^o$}                                                                                                                                  & \multirow{2}{*}{mean} \\ \cline{3-13}
		\multicolumn{2}{c}{Probe}                       & $0^o$            & $18^o$           & $36^o$           & $54^o$           & $72^o$           & $90^o$           & $108^o$          & $126^o$          & $144^o$          & $162^o$          & $180^o$          &                       \\ \hline
		\multirow{8}{*}{NM\#5-6} 	& PoseGait \cite{31}              & 55.3          & 69.6            & 73.9          & 75.0          & 68.0          & 68.2          & 71.1          & 72.9            & 76.1          & 70.4          & 55.4            & 68.7                  \\
		& CNN-LB \cite{5}            & 82.6          & 90.3          & 96.1          & 94.3          & 90.1          & 87.4          & 87.4          & 94            & 94.7          & 91.3          & 78.5          & 89.9                  \\
		& GaitSet \cite{9}              & 90.8          & 97.9          & 99.4          & 96.9          & 93.6          & 91.7          & 91.7          & 97.8          & 98.9          & 96.8          & 85.5          & 95                    \\
		& GaitPart  \cite{10}           & 94.1          & 98.6          & 99.3          & 98.5          & 94            & 92.3          & 92.3          & 98.4          & 99.2          & 97.8          & 90.4          & 96.2                  \\
		& GaitGL \cite{68}                & 96            & 98.3          & 99            & 97.9          & 96.9          & 95.4          & 95.4          & 98.9          & 99.3          & 98.8          & 94            & 97.4                  \\
		& CSTL \cite{66}                 & 97.2          & 99            & 99.2          & 98.1          & 96.2          & 95.5          & 95.5          & 98.7          & 99.2          & 98.9          & 96.5          & 97.8                  \\
		& TransGait \cite{76}            & 97.3          & 99.6          & 99.7          & 99.0          & 97.1          & 95.4          & 97.4          & 99.1          & 99.6          & 98.9          & 95.8          & 98.1                  \\
		& \textbf{MSAFF(ours)} & \textbf{99.1} & \textbf{99.4} & \textbf{99.3} & \textbf{99.1} & \textbf{98.9} & \textbf{98.9} & \textbf{98.9} & \textbf{99.2} & \textbf{99.7} & \textbf{99.6} & \textbf{97.8} & \textbf{99.1}         \\ \hline
		\multirow{8}{*}{BG\#1-2} 	& PoseGait \cite{31}            & 35.3            & 47.2          & 52.4          & 46.9          & 45.5          & 43.9          & 46.1          & 48.1            & 49.4          & 43.6          & 31.1          & 44.5                  \\
		& CNN-LB \cite{5}               & 64.2          & 80.6          & 82.7          & 76.9          & 64.8          & 63.1          & 68            & 76.9          & 82.2          & 75.4          & 61.3          & 72.4                  \\
		& GaitSet \cite{9}             & 83.8          & 91.2          & 91.8          & 88.8          & 83.8          & 81            & 84.1          & 90            & 92.2          & 94.4          & 79            & 87.2                  \\
		& GaitPart \cite{10}             & 89.1          & 94.8          & 96.7          & 95.1          & 88.3          & 94.9          & 89            & 93.5          & 96.1          & 93.8          & 85.8          & 91.5                  \\
		& GaitGL \cite{68}               & 92.6          & 96.6          & 96.8          & 95.5          & 93.5          & 89.3          & 92.2          & 96.5          & 98.2          & 96.9          & 91.5          & 94.5                  \\
		& CSTL \cite{66}                  & 91.7          & 96.5          & 97            & 95.4          & 90.9          & 88            & 91.5          & 95.8          & 97            & 95.5          & 90.3          & 93.6                  \\
		& TransGait \cite{76}            & 94.0          & 97.1          & 96.5          & 96.0          & 93.5          & 91.5          & 93.6          & 95.9          & 97.2          & 97.1          & 91.6          & 94.9                  \\
		& \textbf{MSAFF(ours)} & \textbf{97.7} & \textbf{98.5} & \textbf{98.6} & \textbf{98}   & \textbf{96.9} & \textbf{95.3} & \textbf{96.2} & \textbf{97.6} & \textbf{98.5} & \textbf{97.7} & \textbf{94.1} & \textbf{97.1}         \\ \hline
		\multirow{8}{*}{CL\#1-2} 		& PoseGait \cite{31}             & 24.3          & 29.7          & 41.3          & 38.8          & 38.2            & 38.5          & 41.6          & 44.9          & 42.2          & 33.4          & 22.5          & 36                  \\ 
		& CNN-LB \cite{5}               & 37.7          & 57.2          & 66.6          & 61.1          & 55.2          & 54.6          & 55.2          & 59.1          & 58.9          & 48.8          & 39.4          & 54                    \\
		& GaitSet \cite{9}            & 61.4          & 75.4          & 80.7          & 77.3          & 72.1          & 70.1          & 71.5          & 73.5          & 73.5          & 68.4          & 50            & 70.4                  \\
		& GaitPart \cite{10}            & 70.7          & 85.5          & 86.9          & 83.3          & 77.1          & 72.5          & 76.9          & 82.2          & 83.8          & 80.2          & 66.5          & 78.7                  \\
		& GaitGL \cite{68}               & 76.6          & 90            & 90.3          & 87.1          & 84.5          & 79            & 84.1          & 87            & 87.3          & 84.4          & 69.5          & 83.6                  \\
		& CSTL \cite{66}                 & 78.1          & 89.4          & 91.6          & 86.6          & 82.1          & 79.9          & 81.8          & 86.3          & 88.7          & 86.6          & 75.3          & 84.2                  \\
		& TransGait \cite{76}            & 80.1          & 89.3          & 91.0          & 89.1          & 84.7          & 83.3          & 85.6          & 87.5          & 88.2          & 88.8          & 76.6          & 85.8                  \\
		& \textbf{MSAFF(ours)} & \textbf{92.1} & \textbf{94.6} & \textbf{95.6} & \textbf{93.8} & \textbf{91}   & \textbf{90.6} & \textbf{92.5} & \textbf{94}   & \textbf{95.3} & \textbf{94.8} & \textbf{91.7} & \textbf{93.3}         \\ \hline
	\end{tabular}
	\vspace{-2pt}
\end{table*}
\subsection{Implementation Details}
\textbf{Hyper-parameters.} The structure of the 3-layer graph convolution transformer is shown in \textbf{supplementary material}. For CASIA-B, the output channels of each layer in 4-layer CNN are set to 32, 64, 128, and 128. For Gait3D and GREW, we replace the 4-layer CNN with a 6-layer CNN, and the output channels of each layer are set to 32, 32, 64, 64, 128, and 128. Moreover, a max pooling layer with a stride of 2 is appended after the second convolution layer. In addition, the Leaky ReLU function is applied after each convolutional layer. The parameter $Out_c = 256$. $Z=17$, which is the number of skeleton joints. For CASIA-B, $K = 32$. For Gait3D and GREW, $K = 16$. $N = 30$ when only unimodal inputs are available, and $N = 60$ when bimodal inputs are available. Besides, for CASIA-B, Gait3D, and GREW, the output feature dimensions in the training phase are $339 \times 256$, $195 \times 256$, $195 \times 256$; and in the inference phase are 339, 195, 195.

\textbf{Training Details.} 1) All silhouettes are transformed to 64$\times$44 using the method in \cite{26}. HRNet \cite{65} is applied in CASIA-B for pose estimation. The silhouettes and skeletons are frame-to-frame. 2) All (BA+) triplet loss \cite{28} is employed to train our network. The batch size for training is noted as ($p$, $k$), where $p$ denotes the number of sampled subjects and $k$ denotes the number of sampled sequences for each subject. Particularly, ($p$, $k$) is set to (8, 8) on CASIA-B and (32, 4) on Gait3D and GREW. 3) At the training phase, randomly extract $N$ sorted frames for training. During the testing phase, all frames are used. 4) For the CASIA-B and Gait3D, the model is trained for 100K iterations with the initial Learning (LR) 1e-4, and the LR is multiplied by 0.1 at the 30K and 60K iterations. For the GREW, the model is trained for 210K iterations with the initial LR 1e-4, and the LR is multiplied by 0.1 at the 150K iterations. 5) All models are implemented with PyTorch and trained in RTX 3090 GPU.

\subsection{Comparison with the State-of-the-art Methods}
\textbf{CASIA-B \cite{25}.} To demonstrate the superior performance of our method, it is compared with state-of-art methods on the CASIA-B dataset. The experimental results are illustrated in Table \ref{tbl2}. We draw the following conclusions: 1) MASFF achieves better results under all conditions, which shows that MASFF extracts more robust gait features. 2) Under more difficult conditions, e.g.,  CL, the performance of MASFF is better than GaitSet \cite{9} by 22.9\%, better than GaitPart \cite{10} by 14.6\%, better than GaitGL \cite{68} by 9.7\%, better than CSTL \cite{66} by 9.1\%, better than the multimodal method TransGait \cite{76} by 7.5\%. These suggest that MASFF is more effective in complex environments. 3) MASFF achieves better results as shown in Table \ref{tbl2}; this is due to MSFFS, AFFM, and MSSTFE proposed by us. We will discuss their effects further in the ablation experiments.

\begin{table}[ht]
	\centering
	\caption{Rank-1 accuracy (\%), Rank-5 accuracy (\%), mAP (\%), and mINP on the
		Gait3D dataset. The input size of silhouette is $64 \times 44$. \footnotesize}
	\label{tbl3}
	\resizebox{1.0\columnwidth}{!}{
	\begin{tabular}{ccccc}
		\hline
		Methods   & Rank-1        & Rank-5        & mAP            & mINP                      \\ \hline 
		GEINet \cite{17}    & 5.4           & 14.2          & 5.06           & 3.14                      \\ 
		GaitGraph \cite{66} & 6.25          & 16.23         & 5.18           & 2.42                      \\ 
		GaitPart \cite{10}  & 28.2          & 47.6          & 21.58          & 12.36                     \\ 
		GaitGL \cite{68}    & 29.7          & 48.5          & 22.29          & 13.26                     \\ 
		GLN \cite{72}       & 31.4          & 52.9          & 24.74          & 13.58                     \\ 
		GaitSet \cite{9}   & 36.7          & 58.3          & 30.01          & 17.3                      \\ 
		SMPLGait \cite{75}  & 46.3          & 64.5          & 37.16          & 22.23                     \\ 
		MSAFF (ours)      & \textbf{48.1} & \textbf{66.6} & \textbf{38.45} & \textbf{23.49}            \\ \hline
	\end{tabular} 
}
	\vspace{-2pt}
\end{table}
\begin{table}[ht]
	\centering
	\caption{Rank-1 accuracy ($\%$), Rank-5 accuracy ($\%$), Rank-10 accuracy ($\%$), and Rank-20 accuracy ($\%$) on the GREW dataset.\footnotesize}
	\label{tbl4}
	\resizebox{1.0\columnwidth}{!}{
	\begin{tabular}{ccccc}
		\hline
		Methods  & Rank-1        & Rank-5        & Rank-10       & Rank-20 \\ \hline 
		PoseGait \cite{31}  & 0.23 & 1.05 & 2.23 & 4.28 \\
		GaitGraph \cite{66} & 1.31 & 3.46 & 5.08 & 7.51 \\
		GEINet \cite{17}   & 6.82           & 13.42         & 16.97            & 21.01            \\ 
		TS-CNN \cite{5}  & 13.55          & 24.55          & 30.15          & 37.01            \\ 
		GaitSet \cite{9}  & 46.28          & 63.58          & 70.26          & 76.82          \\ 
		GaitPart \cite{10} & 44.01            & 60.68          & 67.25          & 73.47          \\ 
		TransGait \cite{76}    & 56.27          & 72.72          & 78.12          & 82.51        \\
		MSAFF (ours)      & \textbf{57.40} & \textbf{72.99} & \textbf{78.27} & \textbf{82.88} \\ \hline 
	\end{tabular}
}
	\vspace{-6pt}
\end{table}

\textbf{Gait3D \cite{75}.} Since the CASIA-B dataset is collected in an experimental scenario, we choose to validate the network in the wild scenario on the Gait3D dataset. The experimental results are illustrated in Table \ref{tbl3}. It can be observed that the proposed network achieves the best recognition accuracy. The accuracy of our proposed network is substantially ahead of the two model-based algorithms. It is also competitive for the multimodal algorithm SMPLgait \cite{75}. It can be demonstrated by the results that our proposed network has the same high performance in wild scenarios.

\textbf{GREW \cite{77}.} As GREW has more pedestrians than Gait3D, the network was evaluated on the GREW dataset to verify its generalization and robustness in field situations. The experimental results are illustrated in Table \ref{tbl4}. Our network scores 57.40\% in terms of Rank-1 metric, which exceeds GaitSet and GaitPart by about 10\%. The results demonstrate that our network also performs well on the GREW dataset; This means that our network has learned more useful gait information.

In addition, we find an interesting phenomenon on both Gait3D and GREW: in the wild environment, algorithms using only pose achieve less than 10\% poor results, but when fusing silhouette and pose, pose again makes a sizable contribution to the complete representation of gait. This phenomenon suggests that more modalities allow for a more complete gait representation.

\subsection{Ablation Study}
\textbf{Analysis of FD Pooling.} We analyze the performance of FD Pooling through experiments. As shown in Table \ref{tbl9}: 1) It is possible to reduce the dimension of the gait representations by reducing the $Out_c$, but this approach can significantly reduce the network's performance. 2) When using average pooling, FD Pooling significantly reduces the dimension of the gait representations (the minimum can be 339), and there is almost no loss of accuracy. 3) Accuracy drops more when using other pooling methods. Because when using other pooling methods, only a single dimension is selected to characterize the gait, while using the average pooling method, all dimensions are aggregated to characterize the gait. 4) These experiments prove that FD Pooling is simple and effective.

In the \textbf{supplementary material}, we place ablation studies related to the MSFFS, AFFM, and MSSTFE. Please check.
\begin{table}[]
	\centering
	\caption{Analysis of Feature Dimensional Pooling (FD Pooling) on CASIA-B in terms of averaged rank-1 accuracy. Size means the output feature dimension of network, $Out_c$ means the channel number of FC, and $op$ means the pooling parameter.}
	\label{tbl9}
	\resizebox{1.0\columnwidth}{!}{
	\begin{tabular}{cccccccc}
		\hline
		Size             & $Out_c$ & $Pooling$  & $op$      & NM & BG & CL & Mean \\ \hline
		$339 \times 256$  & 256     & \textbackslash   & \textbackslash  &\textbf{99.1}&\textbf{97.1}&\textbf{93.3}&\textbf{96.5}\\ 
		$339 \times 8$    & 8       & \textbackslash   & \textbackslash  &  97.9  &  93.8  &  90  &93.9      \\ 
		$339 \times 4$    & 4       & \textbackslash   & \textbackslash  & 97   & 93.3   & 85.9    & 92.0     \\ 
		$339 \times 1$    & 1       & \textbackslash   & \textbackslash  & 71.9 & 65.3    & 56.1   & 64.4     \\ 
		$339 \times 8$    & 256     &  Average    &    8      & 98.9   &96.9    &92.9    & 96.2      \\ 
		$339 \times 4$    & 256     &  Average    &    4      & 98.8   &96.8    &92.7    & 96.1      \\ 
		$339 \times 2$    & 256     &  Average    &    2      & 98.8   &96.7    &92.7    & 96.1     \\ 
		$339 \times 1$    & 256     &  Average    &    1      &\textbf{98.7}&\textbf{96.6}&\textbf{92.2}&\textbf{95.9}\\ 
		$339 \times 1$    & 256     &  Max       &    1      & 96.2   &93.1   &88.4    & 92.5     \\ 
		$339 \times 1$    & 256     &  Min       &    1      & 96.1   &93.2    &88.8    & 92.7     \\ 
		$339 \times 1$    & 256     &  Median       &    1      & 97.2   & 92.7   & 85.2   & 91.7     \\
		$339 \times 1$    & 256     &  Average + Max &    1      & 96.1   & 93.1   & 88.4   & 92.5      \\ 
		$339 \times 1$    & 256     &  Average + Min &    1      & 96.1   & 93.2   & 88.9   & 92.7     \\ 
		$339 \times 1$    & 256     &  Average + Median &    1      & 97.7   & 94.2   & 87.2   & 93.0     \\ \hline
	\end{tabular}
}
	\vspace{-8pt}
\end{table}
\section{Conclusion and Future Works}
\label{sec5}
\textbf{Conclusion.} This paper proposes the multi-stage feature fusion strategy and adaptive modal fusion module to exploit the complementary advantages of silhouettes or skeletons. We design the multiscale spatial-temporal feature extractor, which extracts the spatial-temporal linkage feature on different spatial scales to achieve strong spatial-temporal modeling ability. In conjunction with the above, we propose a multi-stage adaptive feature fusion neural network for multimodal gait recognition. Besides, We proposed feature dimensional pooling, which can significantly reduce the dimension of the gait representations without hindering accuracy. Many experiments on CASIA-B, Gait3D, and GREW datasets demonstrate the effectiveness of our proposed network.

\textbf{Future Works.} This paper finds that multimodal algorithms produce better results than unimodal algorithms, suggesting that it may be difficult to represent gait using only a single modal fully. Much information in the raw data and other unused modals has not yet been used, so multimodality may be the next promising research direction for gait recognition.

{\small
\bibliographystyle{ieee}
\bibliography{egbib}
}

\end{document}